\documentclass[times, review, 10pt]{elsarticle}
\usepackage[top=4.3cm, right=4.8cm, bottom=4.3cm, left=4.8cm]{geometry}
\usepackage{multirow} % 如果还没导入，需要在导言区加上
\usepackage{amssymb}
\usepackage{amsmath}
\usepackage{booktabs}
\usepackage{multirow}
\usepackage[table]{xcolor}
\usepackage{makecell}
\usepackage{graphicx}
\usepackage{amsmath,amssymb}
\usepackage[ruled,vlined,linesnumbered]{algorithm2e}
\usepackage{ragged2e}

\newcommand{\Dec}{\mathrm{Dec}}
\newcommand{\clip}{\mathrm{clip}}

\newcolumntype{C}[1]{>{\centering\arraybackslash}m{#1}}
\SetKwInOut{Input}{Input}
\SetKwInOut{Output}{Output}
\SetKw{KwDownTo}{downto}
\SetKw{KwRet}{return}
\journal{Nuclear Physics B}

\begin{document}

\begin{frontmatter}



\title{Flux-Guard: Facial Identity Protection using diffusion models}


\author{Jie Wang}
\author{Tao Wang}
\author{Ru Zhang\corref{cor1}}
\ead{zhangru@bupt.edu.cn}
\author{Jianyi Liu}
\affiliation{organization={School of Cyberspace Security},
            addressline={Beijing University of Posts and Telecommunications}, 
            city={Beijing},
            postcode={100876}, 
            country={China},
}
\affiliation{
            addressline={Nanjing University of Aeronautics and Astronautics}, 
            city={NanJing},
            country={China},
}
\cortext[cor1]{Corresponding author.}
\begin{abstract}
The widespread deployment of face recognition (FR) systems exposes personal images shared on social media and public platforms to identity linkage and privacy risks. Existing adversarial privacy protection methods can degrade unauthorized FR performance but are not compatible with generative face editing. Artificial intelligence-driven face editing tools are gaining popularity, which has significantly increased user demand for personalized portrait generation and social sharing. However, current editing methods often preserve identity features, making the edited images still susceptible to tracking by malicious FR systems. Thus, this paper proposes Flux-Guard, a privacy-preserving face editing framework based on adversarial attacks, which integrates face editing and privacy protection within a unified generative process. Specifically, we design a flow trajectory control method to align semantic manipulations with the generative process and introduce latent-space adversarial optimization with an adaptive perceptual-loss-driven weighting strategy, dynamically adjusting adversarial strength to maximize attack effectiveness while preserving visual quality.

Extensive experiments demonstrate that Flux-Guard supports face editing while significantly improving attack success rates against cross-domain face recognition models on the CelebA-HQ and LADN datasets. Furthermore, evaluation results for commercial APIs have confirmed its effectiveness in real-world applications. The code is released at https://github.com/JLMWang/Flux-Guard.
\end{abstract}


%% Keywords

\end{frontmatter}



\section{Introduction}
\label{sec1}
%% Labels are used to cross-reference an item using \ref command.
In recent years, face recognition technology based on deep neural networks has evolved rapidly and has been deployed on a large scale in various scenarios, including security, financial payments, and access control. While this has brought significant convenience, it has also raised increasingly prominent concerns regarding personal privacy risks \cite{ref1,ref2,ref3,ref4}. Driven in particular by social media, billions of users publicly share their personal images, enabling unauthorized FR systems to associate identities and track behavior across platforms and over time, thereby creating irreversible privacy leakage risks. Thus, it is crucial to develop facial privacy protection technologies to guard against unauthorized black-box recognition and association. 

\begin{figure}[t]
    \centering
    \includegraphics[width=\columnwidth, trim=1.8cm 5.5cm 11.5cm 6.5cm, clip]{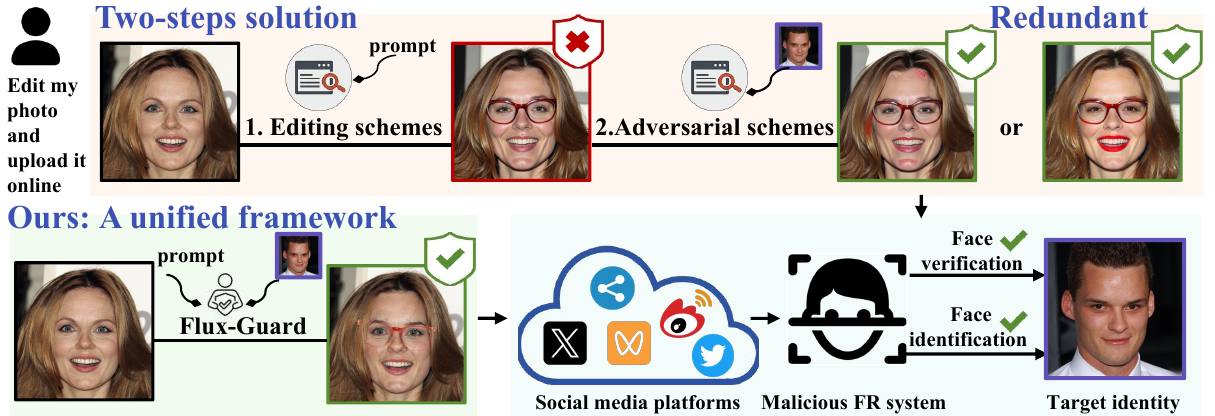}
    \caption{Illustration of the proposed facial privacy protection approach, which integrates face editing and the generation of natural adversarial face images impersonating a specific target identity within a unified pipeline, to deceive malicious  FR systems in both face verification and identification scenarios.}
    \label{fig1}
\end{figure}

Early research commonly superimposed subtle adversarial noise onto original facial images to deceive recognition models. However, while these methods can reduce recognition accuracy, they often result in visible artifacts or texture distortion, thereby compromising the visual quality and social usability of the protected images \cite{ref5,ref6,ref7,ref8,ref9}. In practical applications, an ideal identity protection solution should maintain a natural appearance or make modifications that are barely perceptible to the human eye, thereby striking a balance between privacy protection and visual authenticity. Consequently, recent studies have attempted to achieve camouflage-based privacy protection through makeup transfer, generating appearance changes with reasonable semantic meaning to replace pure noise perturbation \cite{ref10,ref11,ref12,ref13,ref14,ref15, ref16}. However, the cross-model transferability of such methods in black-box scenarios remains limited, and certain makeup styles or patterns are highly visually salient, easily drawing human attention and thus limiting their practicality. Meanwhile, existing research has demonstrated that diffusion models can enhance adversarial capabilities \cite{ref17,ref18,ref19} and generate more effective adversarial samples for popular computer vision tasks \cite{ref20, ref21}. Thus, several attack strategies based on the latent space of diffusion models have emerged. They leverage the strong expressive capability of generative models to perform subtle semantic modifications, thereby significantly improving the visual quality \cite{ref22, ref23, ref24, ref25, ref26,ref27, ref28,ref29,ref30}. However, in pursuing stronger transferability, they may still introduce noticeable artifacts or unnatural local alterations. In summary, research into adversarial examples for facial privacy protection is rapidly advancing in a direction that balances visual naturalness with black-box transferability. It is worth noting that existing studies primarily operate on raw facial images, applying privacy protection directly to the original inputs. With the widespread use of generative editing tools such as diffusion models in content creation and social media dissemination \cite{ref31,ref32,ref33}, users increasingly expect that edited outputs also preserve identity privacy. In response to this practical need, this paper proposes a facial privacy protection framework. This framework is designed for personalized facial editing scenarios and enables both editing and privacy protection within a unified pipeline, aiming to reduce the risk of unauthorized recognition while maintaining editing semantic consistency and visual naturalness, as shown in Fig.\ref{fig1}.

This framework builds on the Rectified-Flow \cite{ref34} diffusion model to enable text-driven editing and incorporates adaptive adversarial optimization in the latent space to generate natural-looking protected images. Specifically, we encode the input image into a latent space and perform an inversion, using the original generation trajectory and attention values within the Diffusion Transformer (DiT) blocks as reusable prior information. Conditioned on the target text prompt, Flux-Guard uses the generative characteristics of flow matching and guides the denoising trajectory to deviate from its original path toward the desired semantic direction by conditionally modulating the velocity field, thereby achieving editing. Here, text-semantic location is employed to generate an editing region mask, and during the generation process, region-specific trajectory control is applied: the editing region adopts text-guided trajectories to inject the target semantics, while non-editing regions follow inverted trajectories to maintain the original regional consistency. Furthermore, to improve the alignment between the edited region and the original facial structure, we employ an editing strategy that separates structure from detail, gradually introducing partial fine-tuning while maintaining structural constraints established in the early stages. In privacy adversarial attacks, we focus on the post-generation stage of the editing process, where detail synthesis plays a central role. We perform iterative adversarial optimization on latent vectors and introduce an adaptive adversarial weighting strategy driven by perceptual loss, which dynamically adjusts the attack intensity within an acceptable range. This approach enhances the transferability of attacks against black-box FR models while preserving visual naturalness.
The main contributions of this paper are as follows:

1)We propose a novel privacy-preserving framework for text-driven face editing, which simultaneously achieves editing and identity adversarial protection within a unified generative pipeline. 

2)We introduce an editing strategy that disentangles structural manipulation from detail refinement, and further introduce mask-guided velocity field fusion to ensure semantically consistent face editing.

3)We design an adaptive adversarial weighting strategy guided by perceptual loss, which dynamically modulates the adversarial strength during latent-space iterative optimization to enhance black-box transferability while maintaining visual quality.

4)Extensive experiments demonstrate that Flux-Guard generates visually natural protected images and achieves competitive protection performance compared with the state-of-the-art methods.

%% Use \subsection commands to start a subsection.
\section{Related works}
\subsection{Noise-Based adversarial attacks}
\label{subsec1}
Early schemes primarily focused on adding noise perturbations within the pixel space to deceive the malicious FR system. Goodfellow et al. \cite{ref5} suggested that the linear behavior of neural networks in high-dimensional spaces can lead to the creation of adversarial examples. Thus, they proposed the Fast Gradient Sign (FGSM) method, which generates adversarial examples by applying minor perturbations to the input data. This method provides a simple and efficient approach for generating adversarial examples. Madry et al.\cite{ref6} employed a PGD approximation with random restarts to approximate the innermost strongest perturbation, and optimized model parameters through adversarial training. Dong et al.\cite{ref7} proposed the Translation-Invariant Attack Method, which optimizes perturbations over multiple translated variants of the input image rather than a single image. This improves transferability in black-box settings and attack success rates. Yang et al.\cite{ref8} proposed an encryption-based facial privacy protection method for FR systems that generates adversarial identity masks. The method conceals identity information by overlaying optimized adversarial masks on facial images, producing protected images that remain visually similar to the originals while effectively deceiving face recognition systems. Zhou et al.\cite{ref9} modeled adversarial examples as noise perturbations superimposed on original facial images that satisfy norm constraints. They designed a joint optimization objective to balance identity deception and evasion of liveness detection, achieving high success rates while maintaining the stealthiness. Research on adversarial examples based on pixel-level noise perturbations includes both classical norm-bounded perturbation methods and more advanced optimization-based methods that generate more effective and transferable perturbations. However, these methods may produce unnatural high-frequency noise that is easily detectable by human observers or defense algorithms.
\subsection{Makeup-Based Adversarial Attacks}
 With the advancement of generative models, more imperceptible attacks have emerged, such as adversarial makeup methods. These schemes leverage generative models to create realistic appearances, significantly enhancing their stealth and cross-model effectiveness. Yin et al. \cite{ref10} proposed using digital makeup to achieve adversarial attacks against FR systems. They applied perturbations to localized regions by generating realistic makeup effects such as eye shadow, ensuring these alterations remained imperceptible to human observers. This method effectively reduced recognition accuracy in black-box models and demonstrated strong transferability. Hu et al. \cite{ref11} proposed an adversarial makeup transfer network that achieves higher attack success rates by jointly optimizing makeup transfer from reference images and adversarial noise. This scheme incorporates a regularization module to balance visual quality and attack strength. Shamshad et al. \cite{ref12} proposed a two-step scheme that finds adversarial latent encodings within the low-dimensional manifold of pre-trained generative models. It utilizes text prompts and identity-preserving regularization to guide search-based adversarial encoding, achieving better transferability and controllability. Shamshad et al. \cite{ref13} employed a randomly initialized neural network architecture combined with a robust correspondence module and a conditional makeup layer to achieve adversarial attacks. Through optimization adjustments during testing, ensure that the generated protected samples maintain the source identity while adopting the makeup style of the reference image, thereby evading detection. Li et al. \cite{ref14} optimized the latent space by introducing global adversarial latent search, enabling the generation of natural and adversarial images without additional information. They employed landmark regularization to ensure generated images preserve original facial features while maintaining privacy. Lyu et al. \cite{ref15} employed a makeup transfer generator in the UV space, combining a makeup adjustment module with a makeup transfer module. They enhanced makeup realism through makeup loss while boosting black-box transferability via an adversarial attack mechanism trained in parallel, thereby generating 3D-oriented protected images. Unlike the aforementioned methods, Sun et al. \cite{ref16} achieved better privacy protection capabilities by incorporating a diffusion model to introduce a makeup removal module and an adversarial makeup transfer module. Using text prompts to establish correspondences between makeup and non-makeup domains, they generated natural adversarial makeup effects. However, such methods still exhibit limited cross-model transferability in black-box scenarios, and certain makeup styles or patterns retain pronounced visual saliency.
 \subsection{Diffusion Model-based semantic invariant adversarial attacks}
An et al. \cite{ref23} utilized text embeddings of target images as conditional guidance. During DDIM inversion and denoising sampling, they employed h-space features from the UNet encoder as adversarial optimization signals to generate protected images. This enables better black-box privacy protection without relying on additional classification models. Hu et al. \cite{ref24} proposed alternately injecting adversarial gradient perturbations and residual constraints during the denoising process. This gradually shifts the benign distribution toward the adversarial distribution, thereby generating adversarial facial samples with higher transferability and improved perceptual quality. Liu et al. \cite{ref25}  encoded input face images into semantic latent variables using a diffusion autoencoder. They then performed iterative adversarial optimization on these semantic codes within the latent space while incorporating semantic consistency constraints. This scheme generated natural-looking protected images that effectively deceived facial recognition systems. Wang et al. \cite{ref26} proposed iteratively injecting gradient-based adversarial identity guidance during the denoising process. Combined with structure-preserving regularization and a temporal step truncation strategy, this achieved target identity impersonation with natural appearance, generating high-quality adversarial images. Han et al. \cite{ref27} applied pre-trained diffusion models to perform dual-face guidance during sampling, preserving the visual identity information of the source image while extracting generic embeddings from the target face. They manipulated the denoising trajectory through adversarial attacks, rendering the alterations imperceptible to human observers. Li et al. \cite{ref28} proposed latent space and DDIM inversion with denoising iterations to optimize along adversarial directions on latent variables. They constrained modification magnitude using self-attention structure loss and perceptual loss, while enhancing black-box transferability through facial attribute embedding alignment. Salar et al. \cite{ref29} proposed learning step-by-step unconditional embeddings as null-text guides to preserve identity-related adversarial modifications, performing adversarial optimization directly on latent variables. Simultaneously, they employed self-attention structure consistency constraints to maintain human-recognizable structural integrity. Li et al. \cite{ref30} jointly optimized adversarial representations in semantic latent spaces and diffusion latent spaces. They employed adaptive optimization and latent variable regularization constraints to maximize adversarial gradient utilization while preserving natural facial appearance and source image consistency, thereby generating protected facial images.
 \subsection{Text-to-Image diffusion models}
 Advances in text-to-image diffusion models \cite{ref35,ref36} have significantly propelled recent progress in text-conditioned image and video generation \cite{ref37,ref38,ref39,ref40}. Recent improvements have primarily manifested in enhanced generation quality and better text alignment capabilities. SDXL \cite{ref41} achieves greater stability in detail representation and overall consistency across complex scenes by expanding model capacity and employing multi-encoder conditional modeling with a two-stage generation process. Subsequently, research began incorporating Transformers as diffusion backbones to enhance cross-modal interaction and long-text modeling. Stable Diffusion 3 \cite{ref42} employs multimodal Transformers, achieving further improvements in long-text comprehension and photorealistic generation. Aligning with this trend, FLUX adopts a Transformer-based latent generative framework and leverages Rectified Flow to formulate sampling as ODE-based flow integration. This design further enhances text alignment and image quality, establishing itself as a representative open-source foundation model in recent years \cite{ref43}. 
\section{Flux-Guard}
\subsection{Preliminaries}
\subsubsection{FLUX model and Rectified Flow}
In recent years, diffusion-based text-to-image generation models have increasingly shifted from UNet backbones to Transformer-based architectures, with the Flux model \cite{ref43} standing as one of the representative open-source foundational models. FLUX.1 is a rectified flow transformer \cite{ref34,ref43,ref44} trained in the latent space of an image autoencoder \cite{ref45}. It replaces the conventional UNet denoising network with a Transformer backbone, performing conditional fusion and representation updates on latent tokens. The Flux model adopts a hybrid architecture composed of double-stream and single-stream blocks \cite{ref46}.

Rectified Flow \cite{ref34} establishes a smooth transition between the Gaussian noise distribution
$\pi_0$ and the real data distribution $\pi_1$ along a straight path. This is realized by learning a forward-simulating system defined by the ODE, which maps $Z_0 \sim \pi_0 \ \text{to} \ Z_1 \sim \pi_1$:
\begin{equation}
dZ_t = v(Z_t, t)\,dt, \qquad t \in [0,1]
\end{equation}

In practice, the velocity field $v(\cdot)$ is parameterized by a neural network $v_{\theta}(Z_t, t)$.
Rectified Flow defines the forward noising process as a linear function with time $t \in [0,1]$:
\begin{equation}
Z_t = tZ_1 + (1-t)Z_0 \;\Rightarrow\; dZ_t = (Z_1 - Z_0)\,dt.
\end{equation}

Subsequently, $v_{\theta}(Z_t, t)$ can be optimized via the method of least squares regression:
\begin{equation}
\min_{\theta} \int_{0}^{1} \mathbb{E}\left[ \left\| (Z_1 - Z_0) - v_{\theta}(Z_t, t) \right\|^2 \right] dt.
\end{equation}

Given $N$ discrete timesteps $t = \{t_N, \ldots, t_0\}$, the model iteratively predicts $v_{\theta}(Z_{t_i}, t_i)$ for $i = N, \ldots, 1$, then backward process is defined by the following function:
\begin{equation}
Z_{t_{i-1}} = Z_{t_i} + (t_{i-1} - t_i)\, v_{\theta}(Z_{t_i}, t_i).
\end{equation}

Rectified Flow defines a straightforward transformation path from Gaussian noise to real images, enabling high-quality image generation in fewer steps than DDPM \cite{ref47}.
\subsubsection{Adversarial problem definition}
When exposed to malicious FR systems, a face image $I$ is primarily subject to two privacy leakage risks: verification and identification. For verification, the FR system maps facial images into a feature space using a feature extractor, then determines whether two images correspond to the same identity by comparing their feature similarity with a threshold $\tau$. For identification, the FR system compares a probe image against candidate identities in image database $G$, returning the identity with the highest similarity as the recognition result. In practice, online social platforms and third-party services often deploy FR models as black-box systems. Consequently, once users upload raw facial images, unauthorized parties may perform verification or identification, leading to privacy risks such as identity association and cross-platform tracking. Thus, facial privacy protection schemes are required to deceive the system in both modes, ensuring that protected images maintain natural visual appearance and human-perceivable identity consistency while preventing unauthorized systems from correctly verifying or identifying them. Specifically, suppose that the FR model $f$ encodes a face image $I$ into a feature vector $f(I) \in \mathbb{R}^d$. $d$ indicates the feature dimension, $\mathrm{Sim}(\cdot)$ denotes feature similarity, $\tau$ represents the system decision threshold.

Given a source image $I_{\mathrm{src}}$ and a target image $I_{\mathrm{tgt}}$, our objective is to generate a protected image $I_{\mathrm{adv}}$ that preserves visual naturalness, while being closer to the target identity than to the source identity in the feature space, thereby enabling targeted impersonation and reducing true-identity exposure. For face verification scenarios, $\mathrm{Sim}\!\left(f(I_{\mathrm{adv}}),\, f(I_{\mathrm{tgt}})\right) \geq \tau$ should be satisfied to ensure that $I_{\mathrm{adv}}$ is incorrectly identified as the target identity during verification. For face identification scenarios, the system returns the identity in the image database that most closely matches the probe image. To ensure the target identity is the optimal match, it must satisfy:
\begin{equation}
\mathrm{Sim}\!\left(f(I_{\mathrm{adv}}),\, f(I_{\mathrm{tgt}})\right)
>\mathrm{Sim}\!\left(f(I_{\mathrm{adv}}),\, f(I_g)\right), \quad \forall\, I_g \in G,\ I_g \neq I_{\mathrm{tgt}} \ 
\end{equation}
Meanwhile, to ensure image usability and visual naturalness, we introduce the constraint term $\mathcal{H}(I_{\mathrm{adv}}, I_{\mathrm{src}}) \leq \varepsilon $.
\subsection{Methodology}
\begin{figure}[htbp]
    \centering
    \includegraphics[width=\linewidth, trim=1.3cm 3.3cm 1.2cm 5.1cm, clip]{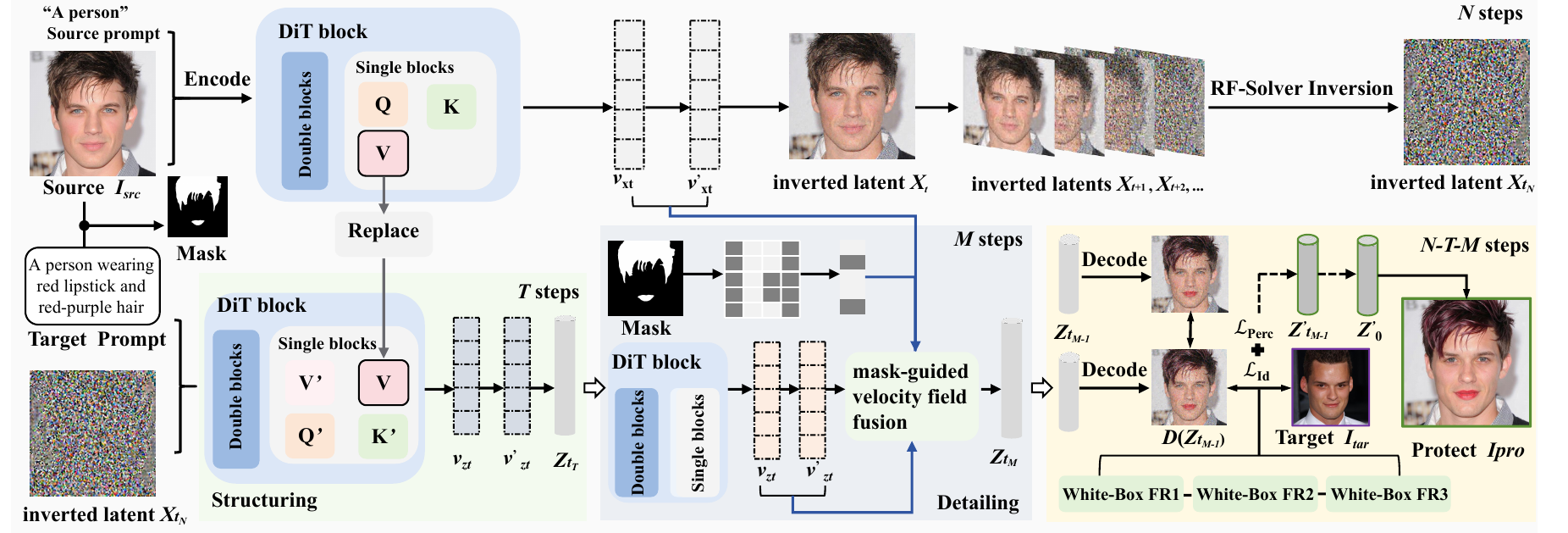}
    \caption{Overview of the proposed Flux-Guard. Flux-Guard first performs value feature $ V$ replacement within $T$ steps to inject the structural information, followed by mask-guided velocity field fusion within $M$ steps to achieve fine-grained editing. Then, Flux-Guard utilizes adversarial guided diffusion to generate natural adversarial face images.}
    \label{fig2}
\end{figure}

Our goal is to protect facial privacy against unauthorized recognition systems while enabling controllable, prompt-driven generative editing, thereby providing a unified framework that balances privacy preservation and editability. The overall pipeline of Flux-Guard is shown in Fig.\ref{fig2}. Flux-Guard replaces attention values within the Diffusion Transformer (DiT) module to inject global structural information, subsequently introducing mask-guided flow-trajectory control for attribute editing to achieve better customized manipulation. It further applies iterative adversarial optimization with a perceptual-loss-guided adaptive weight schedule, enabling the edited images to evade FR systems while preserving visual consistency for privacy protection.
\subsubsection{Prompt-guided face editing}
To ensure natural and prompt-aligned face editing, we introduce a structural injection strategy in the early denoising stage. Inspired by RF-Edit[32], we cache the attention value features $\{V_{t_n,l}\}_{l=1}^{q}$ and $\{V_{t_{n+\Delta t_i},l}\}_{l=1}^{q}$
from the last $q$ Single-Stream Transformer blocks at each timestep during inversion. This prior information facilitates the basic preservation of source structures during the generation process. Under prompt guidance, the diffusion model progressively converges toward textual semantic information during the denoising process, and generates attention value features $\{\tilde{V}_{t_n,l}\}_{l=1}^{q}$ and $\{\tilde{V}_{t_{n+\Delta t_i},l}\}_{l=1}^{q}$. For the first $T$ denoising steps, we replace the generated values with the cached ones when computing attention:
\begin{equation}
\tilde{F}_{t_n}^{\,l} = \operatorname{Attention}\!\left(\tilde{Q}_{t_n}^{\,l},\, \tilde{K}_{t_n}^{\,l},\, V_{t_n}^{\,l}\right)
\end{equation}
\begin{equation}
\tilde{F}_{t_{n+\Delta t_i}}^{\,l}
= \operatorname{Attention}\!\left(\tilde{Q}_{t_{n+\Delta t_i}}^{\,l},\, \tilde{K}_{t_{n+\Delta t_i}}^{\,l},\, V_{t_{n+\Delta t_i}}^{\,l}\right)
\end{equation}
Where $F$ denotes the output feature of the self-attention module and $Q$, $K$, $V$ represent query, key, and value for attention, respectively. This injection provides stable structural constraints from the source face, preserving global configuration and geometric consistency. Subsequently, we introduce a detailed editing strategy guided by flow trajectories with masking constraints to achieve natural face editing. 

During denoising, a standard reconstruction follows a stable denoising trajectory guided by the source prompt. Specifically, we conditionally modify the velocity field in target regions via target prompts to align the denoising trajectory with the text, while unedited regions retain their initial trajectory. This differentiated processing enables greater controllability in detail editing. To formalize this, let $\{X_{t_n}\}_{n=0}^{N}$ denote the latent sequence obtained by inversion with the source prompt, and let $\{Z_{t_n}\}_{n=0}^{N}$ denote the latent sequence during forward denoising for editing with the target prompt. At each timestep $t_n$, the flow model predicts a velocity field $v_{\theta}(\cdot, t_n, c)$ conditioned on prompt embedding $c$. In inversion, we record the predicted velocities at the two solver evaluation times used by the second-order update, namely:
\begin{equation}
u_{t_n} = v_{\theta}(x_{t_n}, t_n, c_s)
\end{equation}
\begin{equation}
u_{t_n+\Delta t} = v_{\theta}(x_{t_n+\Delta t}, t_n+\Delta t, c_s)
\end{equation}

For editing, we acquire the binary mask $\mathcal{M}$ through prompting and convert it to the latent space, where $\mathcal{M} = 1$ indicates regions to be edited and $\mathcal{M} = 0$ indicates regions to be preserved. At selected denoising steps $n \in \mathcal{N}$, we mix the target-prompt velocity with the cached inversion velocity in a spatially controlled manner.
\begin{equation}
\hat{v}_{t_n}
= \mathcal{M} \cdot v_{\theta}(z_{t_n}, t_n, c_t)
+ (1 - \mathcal{M})\cdot u_{t_n}
\end{equation}
\begin{equation}
z_{t_n+\Delta t}
= z_{t_n} + \Delta t \cdot \hat{v}_{t_n}
\end{equation}
\begin{equation}
\hat{v}_{t_n+\Delta t}
= \mathcal{M} \cdot v_{\theta}(z_{t_n+\Delta t}, t_n+\Delta t,  c_t)
+ (1 - \mathcal{M})\cdot u_{t_n+\Delta t}
\end{equation}

This formulation ensures that the areas marked for editing are guided by the target-prompt velocity, while the unedited regions retain the original inversion-derived velocity. The mixture of velocities in different regions allows for precise control over the target attributes, without compromising the integrity of the original image.
\subsubsection{Adversarial optimization-driven facial protection}
During the late stages of generation, we perform iterative adversarial optimization on the adversarial latent branch to weaken facial recognition matching while retaining visual consistency. Let the adversarial latent be $Z^{\mathrm{adv}}$, its corresponding clean branch latent be $Z^{\mathrm{clean}}$. Adversarial optimization is conducted in the latent space, but its constraints necessitate decoding both solutions into the pixel space to serve as the foundation for both optimization and constraint enforcement: $I^{\mathrm{adv}} = D\!\left(Z^{\mathrm{adv}}\right)$ and $I^{\mathrm{clean}} = D\!\left(Z^{\mathrm{clean}}\right)$,$D(\cdot)$ denotes the VAE decoder.
To enhance the transferability to black-box models, we pre-extract reference features from the target reference image $ I^{\mathrm{tgt}} $ across three FR systems $\mathcal{R}$:
\begin{equation}
e_{r}^{\mathrm{tgt}} = r\!\left(N(I^{\mathrm{tgt}}, s_r)\right), \quad \forall\, r \in \mathcal{R}
\end{equation}
Where $N(\cdot, s_r)$ denotes the preprocessing operation that performs size alignment and normalization on the input to adapt it to the $r$-th recognition model. For the current adversarial image, the features are similarly obtained:
\begin{equation}
e_{r}^{\mathrm{adv}} = r\!\left(N(I^{\mathrm{adv}}, s_r)\right), \quad \forall\, r \in \mathcal{R}
\end{equation}

Subsequently, features from multiple FR models are integrated to construct an identity adversarial loss using cosine similarity as the metric:
\begin{equation}
L_{\mathrm{id}}(n)
=
1 - \frac{1}{|R|}
\sum_{r \in R}
\cos\!\left(e_{r,n}^{\mathrm{adv}},\, e_{r}^{\mathrm{tgt}}\right)
\end{equation}
The smaller the loss, the more similar the adversarial sample is to the target identity, thereby deceiving the recognition system. To suppress perceptible distortion caused by adversarial perturbations, we introduce a perceptual consistency constraint. Let $\phi(\cdot)$ denote the output of the VGG feature extractor at a specified layer. The perceptual loss is defined as:
\begin{equation}
L_{\mathrm{perc}}(n)
=
\frac{1}{P}
\sum_{i=1}^{P}
\left(\phi(I^{\mathrm{adv}})_i - \phi(I^{\mathrm{clean}})_i\right)^2
\end{equation}

Notably, we propose an adaptive adversarial weighting strategy driven by perceptual loss to dynamically adjust the intensity of identity attacks, adapting to different images. Specifically, the current perception distortion metric is quantified as $p_n = L_{\mathrm{perc}}(n)$, let the distortion threshold be $\tau$, the lower and upper bounds of the weight be $(\omega_{\min}, \omega_{\max})$, and adopt a sigmoid scheduling function with slope coefficient $\kappa$.
\begin{equation}
\omega_{\mathrm{id}}(n)
=
\omega_{\min}
+
\frac{\omega_{\max} - \omega_{\min}}
{1 + \exp\!\left(\kappa(p_n - \tau)\right)}
\end{equation}

When $p_n \leq \tau$, then $\exp\!\left(\kappa(p_n - \tau)\right)$ is small and $\omega_{\mathrm{id}}(n)$ tends toward $\omega_{\max}$, thereby enhancing  the identity attack. When $p_n > \tau$, $\omega_{\mathrm{id}}(n)$ rapidly decreases toward $\omega_{\min}$, thereby automatically reducing the adversarial strength to prioritize visual quality. To further ensure stability, the clipped form is explicitly defined as:
\begin{equation}
\omega_{\mathrm{id}}(n)
=
\max\!\left(\omega_{\min},\, \min\!\left(\omega_{\max},\, \omega_{\mathrm{id}}(n)\right)\right)
\end{equation}

Ultimately, during the adversarial optimization process $n$ steps, $m$ inner-loop gradient updates are performed on the latent variable. The objective function for the $n$-th iteration is defined as:
\begin{equation}
L(n,m)
=
\omega_{\mathrm{id}}(n,m)\,L_{\mathrm{id}}(n,m)
+
\lambda\,L_{\mathrm{perc}}(n,m)
\end{equation}

Where $\lambda$ denotes the perceptual loss weighting coefficient, the corresponding update for the latent vector is:
\begin{equation}
Z_n^{(m+1)}
\leftarrow \text{Adam}(Z_n^{(m)}, \nabla_{Z_n^{(m)}} L(n,m))
\end{equation}

This strategy enables adversarial mechanisms to adaptively adjust during training iterations in response to visual distortion: when perturbations have not yet significantly compromised visual coherence, the model prioritizes enhancing protective capabilities; when perceived distortion accumulates near a threshold, weight scheduling automatically suppresses adversarial strength, thereby achieving synergistic optimization of adversarial performance and visual fidelity. The Flux-Guard algorithm is summarized in Algorithm \ref{algorithm1}.
{\setlength{\textfloatsep}{4pt}
\setlength{\floatsep}{4pt}
\setlength{\intextsep}{4pt}
\setlength{\abovecaptionskip}{2pt}
\setlength{\belowcaptionskip}{0pt}
\begin{algorithm}[!t]
\small
\caption{Flux-Guard pipeline with Adversarial Optimization}
\label{algorithm1}
\Input{\justifying Source image $I_{src}$, target image $I_{tgt}$, source prompt $c_s$, target prompt $c_t$, mask $\mathcal{M}$, timesteps $N$, early injection steps $T$, editing steps $M$, perceptual extractor $\phi$, inner steps $m$, adaptive params $\tau,\omega_{\min},\omega_{\max},\kappa$, perceptual weight $\lambda$.}
\Output{Protected edited image $I_{pro}$.}

% Calculate latent codes, cache attention values, and predicted velocities
Calculate benign latent codes $\{X_{t_n}\}_{n=1}^{N}$, consistent noise maps $\{Z_{t_n}\}_{n=1}^{N}$, cache the attention value features ${V}^{cache}_{t_n}$, predicted velocities $u_{t_n}$, $u_{t_n+\Delta t}$

\For{$n \gets 1$ \KwTo $N$}{
  % Early Injection Step (for $n < T$)
  \If{$n \in [1, T)$}{
    $\tilde V^{l}_{t_n} \leftarrow V^{l,\mathrm{cache}}_{t_n}$, \quad $\tilde F^{l}_{t_n} = \mathrm{Attention}(\tilde Q^{l}_{t_n}, \tilde K^{l}_{t_n},  V^{l,\mathrm{cache}}_{t_n})$ \;
    $\tilde V^{l}_{t_n+\Delta t} \leftarrow V^{l,\mathrm{cache}}_{t_n+\Delta t}$, \quad $\tilde F^{l}_{t_n+\Delta t} = \mathrm{Attention}(\tilde Q^{l}_{t_n+\Delta t}, \tilde K^{l}_{t_n+\Delta t}, V^{l,\mathrm{cache}}_{t_n+\Delta t})$\;
  }
  
  % Editing Step (for $T \le k < M$)
  \If{$n \in [T, M)$}{
    $\hat v_{t_n} \leftarrow v_\theta(z_{t_n}, t_n, c_t)$
    $\hat v_{t_n+\Delta t} \leftarrow v_\theta(z_{t_n} + \Delta t\, \hat v_{t_n},\, t_n + \Delta t,\, c_t)$\;
    
    $\hat v_{t_n} \leftarrow \mathcal{M} \cdot \hat v_{t_n} + (1 - \mathcal{M}) \cdot u_{t_n}$\;
    $\hat v_{t_n + \Delta t} \leftarrow \mathcal{M} \cdot \hat v_{t_n + \Delta t} + (1 - \mathcal{M}) \cdot u_{t_n + \Delta t}$\;
    ${Z}_{t_{n+1}} = {Z}_{t_n} + (t_{n+1} - t_n) \hat v_{t_n}+ \frac{1}{2}(t_{n+1} - t_n)^2 \cdot \frac{\hat{v}_{t_n+\Delta t}-\hat{v}_{t_n}}{\Delta t}$\;
  }
  % Adversarial Optimization (for $M \le k < N$)
  \If{$n \in [M, N)$}{
    \For{$m = 0$ \KwTo $m-1$}{
      $I^{adv} \leftarrow \Dec(z^{adv})$, $I^{clean} \leftarrow \Dec(z^{clean})$\;
      $L_{id}(m) = 1 - \frac{1}{|\mathcal{R}|} \sum_{r \in \mathcal{R}} \cos\left(e_r^{\text{adv}}, e_r^{\text{tgt}}\right)$\;
      $L_{\text{perc}}(m) = \frac{1}{P} \sum_{i=1}^{P} \left( \phi(I^{\text{adv}})_i - \phi(I^{\text{clean}})_i \right)^2$\;

      $\omega_{id} \leftarrow \omega_{\min} + \frac{\omega_{\max} - \omega_{\min}}{1 + \exp\left(\kappa(L_{\text{perc}}(m) - \tau)\right)}$\;
      $\omega_{id} \leftarrow \clip(\omega_{id}, \omega_{\min}, \omega_{\max})$\;
      
      $L \leftarrow \omega_{id} L_{id} + \lambda L_{\text{perc}}$\;
      ${Z}_{adv} \leftarrow \text{Adam}({Z}_{adv}, \nabla_{{Z}_{adv}} L)$\;
    }

    $Z_{t_n} \leftarrow Z^{adv}$\;
  }
}
$I_{pro} \leftarrow \Dec(Z_{t_n})$

\KwRet $I_{pro}$
\end{algorithm}
}
\section{Experiments}
\subsection{Experimental settings}
\begin{table}[t]
\caption{Comparison with state-of-the-art methods on CelebA-HQ and LADN datasets for targeted black-box attacks. FaceN: FaceNet, MF: MobileFace, Avg: Average.}
\centering
\setlength{\tabcolsep}{1.0pt}
\renewcommand{\arraystretch}{1.05}
\resizebox{\linewidth}{!}{
\begin{tabular}{ll*{5}{C{1.1cm}}C{0.01cm}*{5}{C{1.1cm}}}
\hline
\multirow{2}{*}{Method}
& \multirow{2}{*}{\makecell[c]{Dataset\\ASR\% $\uparrow$}}
& \multicolumn{5}{c}{CelebA-HQ}
&
& \multicolumn{5}{c}{LADN-dataset} \\
\cline{3-7} \cline{9-13}
&
&
IR152 & IRSE50 & FaceN & MF & Avg
&
&
IR152 & IRSE50 & FaceN & MF & Avg \\
\hline

Clean
& --
& 4.50 & 12.7 & 1.60 & 23.1 & 10.4
&
& 0.90 & 3.30 & 0.60 & 6.30 & 2.77 \\
\hline

\multirow{2}{*}{\makecell[l]{Noise-\\Based}}
& PGD (ICLR'18)
& 20.6 & 36.8 & 1.85 & 43.9 & 25.7
&
& 19.5 & 40.0 & 3.8 & 41.0 & 26.0 \\

& TIP-IM (ICCV'21)
& 50.0 & 72.1 & 11.5 & 79.0 & 53.1
&
& 50.5 & 45.3 & 15.1 & 55.3 & 41.5 \\
\hline

\multirow{3}{*}{\makecell[l]{Makeup\\-Based}}
& AMT-GAN (CVPR'22)
& 35.1 & 76.9 & 16.6 & 50.7 & 44.8
&
& 49.1 & 89.6 & 32.1 & 72.4 & 60.8 \\

& Clip2Protect (CVPR'23)
& 41.8 & 85.7 & 41.6 & 78.7 & 61.9
&
& 41.5 & 68.3 & 42.1 & 50.8 & 50.6 \\

& DiffAM (CVPR'24)
& 69.5 & 93.6 & 53.7 & 87.0 & 75.9
&
& 59.9 & 93.5 & 43.6 & 84.1 & 70.2 \\

\hline

\multirow{3}{*}{\makecell[l]{Facial\\Semantic\\Invariant}}
& Adv-Diffusion (AAAI'24)
& 43.1 & 79.8 & 26.0 & 78.6 & 56.8
&
& 25.9 & 46.6 & 17.5 & 40.6 & 32.6 \\

& EFPP (CVPR'25)
& 71.2 & 90.9 & 60.5 & 90.4 & 78.2
&
& 70.2 & 91.8 & 48.1 & 84.2 & 73.5 \\

& DiffProtect (PR'26)
& 52.5 & 73.9 & 36.8 & 71.1 & 58.5
&
& 33.9 & 52.6 & 23.4 & 47.2 & 39.2 \\

\hline

\rowcolor[rgb]{0.80,0.84,0.93}
Facial edit
& Flux-Guard (Ours)
& 85.4 & 98.5 & 62.9 & 99.5 & 86.5
&
& 73.4 & 94.8 & 49.3 & 84.9 & 75.6 \\
\hline
\end{tabular}
}
\label{table1}
\end{table}

Dataset. We evaluate targeted attacks on both face verification and face identification. For face verification, CelebA-HQ \cite{ref48} and LADN \cite{ref49} are used as test sets for impersonation attacks. From CelebA-HQ, we randomly select 1,000 images and divide them into four groups, each with one target identity from \cite{ref11}. For LADN, its 332 images are also divided into four groups to impersonate the target identities defined in \cite{ref11}. For face identification, we employ the probe set comprising 500 randomly selected images from the CelebA-HQ dataset. The gallery set is constructed using the corresponding 500 images from the same identities, alongside four additional target identities provided.

Baseline Methods. We conduct performance comparisons against multiple adversarial attack baselines, including PGD \cite{ref6}, TIP-IM \cite{ref8}, AMT-GAN \cite{ref11}, Clip2Protect \cite{ref12}, DiffAM \cite{ref16}, Adv-Diffusion \cite{ref22}, DiffProtect \cite{ref25}, EFPP \cite{ref29}. PGD and TIP-IM represent relatively classical noise-based attack methods, while AMT-GAN, Clip2Protect, and DiffAM adopt makeup-based strategies to generate protected images. Meanwhile, Adv-Diffusion, DiffProtect, and EFPP are designed for the diffusion-based generation model, generating protected facial images in a manner that preserves semantic invariance.

Target Models. We employ four widely used FR models under the current task to evaluate protection performance: IR152 \cite{ref50}, IRSE50 \cite{ref51}, FaceNet \cite{ref52}, and MobileFace \cite{ref53}. Three of these models serve as the white-box model set, while the remaining one acts as the black-box model for testing. To further validate real-world effectiveness, we additionally conduct experiments on two commercial face recognition APIs, Face++ and Aliyun FR.

Evaluation Metrics. For face verification, we employ ASR \cite{ref11} (the ratio of successfully attacked adversarial examples to all adversarial examples) to calculate the proportion of successful misclassifications by malicious FR systems. For face identification, Rank-1 and Rank-5 serve as evaluation metrics: an attack is deemed successful if the target identity appears in the top-ranked position or within the top five positions after sorting gallery images by similarity to the probe image. For commercial APIs, we utilize the confidence scores returned by the service, where a higher score indicates greater identity similarity. Furthermore, to evaluate the quality of generated images, we employ FID \cite{ref54}, PSNR(dB), and SSIM as quality metrics.
\begin{table}[t]
\caption{Comparison of attack success rates for black-box attacks on face identification tasks. R1-T: Rank-1-Targeted. R5-T: Rank-5-Targeted.}
\centering
\setlength{\tabcolsep}{6pt}
\renewcommand{\arraystretch}{1.05}
\resizebox{\linewidth}{!}{
\begin{tabular}{lcccccccccc}
\hline
Method & \multicolumn{2}{c}{IR152} & \multicolumn{2}{c}{IRSE50} & \multicolumn{2}{c}{FaceNet} & \multicolumn{2}{c}{MobileFace} & \multicolumn{2}{c}{Average} \\
\cline{2-11}
& R1-T & R5-T & R1-T & R5-T & R1-T & R5-T & R1-T & R5-T & R1-T & R5-T \\
\hline

Clip2Protect (CVPR'23) & 13.4 & 42.5 & 33.6 & 68.8 & 18.6 & 42.4 & 20.2 & 60.0 & 21.4 & 53.4 \\

DiffAM (CVPR'24) & 29.4 & 55.6 & 50.2 & 89.6 & 22.6 & 61.0 & 30.0 & 72.4 & 33.0 & 69.9 \\

Adv-Diffusion(AAAI'24) & 11.2 & 40.0 & 20.6 & 79.4 & 13.7 & 37.2 & 15.8 & 59.2 & 15.3 & 53.9 \\
EFPP(CVPR'2025) & 44.4 & 71.2 & 61.4 & 92.4 & 23.6 & 63.2 & 41.2 & 81.4 & 42.6 & 77.0 \\

DiffProtect (PR'26) & 15.6 & 46.2 & 16.8 & 68.4 & 15.4 & 47.0 & 12.4 & 46.8 & 15.0 & 52.1 \\
\rowcolor[rgb]{0.80,0.84,0.93}
Flux-Guard (Ours) & 46.6 & 82.2 & 78.8 & 98.9 & 26.3 & 67.8 & 82.4 & 95.4 & 58.5 & 86.0 \\
\hline
\end{tabular}
}
\label{table2}
\end{table}
Implementation Details. Flux-Guard adopts FLUX.1-dev \cite{ref43} as the foundation model, with its weights frozen. The total number of inversion steps is set to $N=25$. We use an Adam optimizer in adversarial optimization, with the learning rate set to $1 \times 10^{-2}$. The attention value replacement steps are set to $T=5$, and the mask-constrained flow trajectory editing steps are set to $M=20$. The adversarial optimization step is set to 5, with the internal iteration count $m=10$. The perceptual weight $\lambda$ is set to 0.5, the gradient setting $\kappa$ is set to 20, the threshold $\tau$ is set to 0.3, and $\omega_{\min}$ and $\omega_{\max}$ are set to 0.05 and 1, respectively. All images are resized to $512 \times 512$, and all experiments are conducted on a single NVIDIA RTX A6000 GPU.
\begin{figure}[t]
    \centering
    \includegraphics[width=\columnwidth, trim=1.3cm 2cm 3.5cm 2cm, clip]{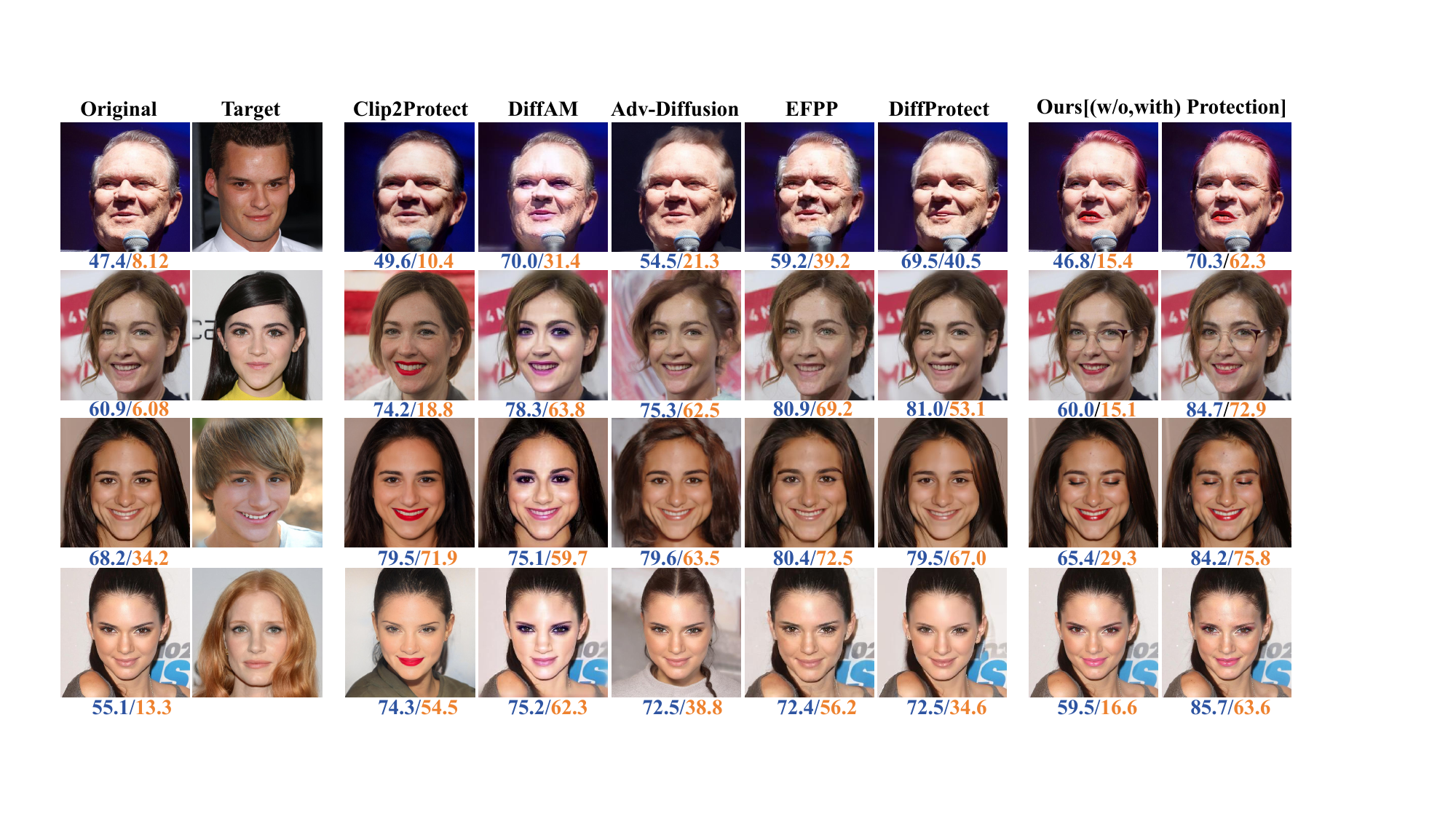}
    \caption{Visual quality comparison of adversarial examples generated by adversarial methods on the CelebA-HQ dataset. The images generated by Flux-Guard implement both face editing and effective facial privacy protection. Blue/yellow (1/2) numbers below each image: confidence scores returned by Face++ and Aliyun (higher is better). First line: purple-red hair and lipstick. Second line: wearing glasses and lipstick. Third line: eyes closed and lipstick. Fourth line: pink eyeshadow and lips.}
    \label{fig3}
\end{figure}

\subsection{Comparison with the state-of-the-art methods}
This section compares the proposed method with state-of-the-art methods in terms of attack performance and image quality using black-box scenarios. Each FR model employed has an error acceptance rate set to 0.01, specifically IR152 (0.167), IRSE50 (0.241), FaceNet (0.409), and MobileFace (0.302). Face verification scenarios. Table \ref{table1} presents the ASR results for black-box attacks against four FR models on the CelebA-HQ and LADN datasets. We evaluated the performance of impersonation attacks using four target identities. As shown in the table, Flux-Guard achieves an average ASR of 86.5\% and 75.6\%, demonstrating its strong effectiveness in generating adversarial samples. Compared to SOTA methods, our scheme achieves higher ASR results across different FR models, indicating better transferability while maintaining robust attack performance.
Face identification scenarios. Table \ref{table2} presents Rank-1 and Rank-5 results on the CelebA-HQ dataset, evaluating its impersonation attack capability. The table shows our method achieved average ASR rates of 58.5\% and 86.0\% for Rank-1 and Rank-5 attacks, respectively, surpassing the best SOTA approach by 15.9\% and 9\%. This demonstrates that our method possesses robust attack capabilities across various black-box models.
\begin{table}[!b]
\caption{Comparison of different methods on image quality metrics. Arrows indicate whether higher (↑) or lower (↓) values are better.}
\centering
\footnotesize
\setlength{\tabcolsep}{11pt}
\renewcommand{\arraystretch}{1.05}
\begin{tabular}{lccc}
\hline
Method & FID $\downarrow$ & SSIM $\uparrow$ & PSNR $\uparrow$ \\
\hline
Clip2Protect (CVPR'23) & 28.11 & 0.5537 & 18.92 \\
DiffAM (CVPR'24) & 27.42 & 0.5912 & 18.81 \\
Adv-Diffusion(AAAI'24) & 30.51 & 0.5123 & 14.85 \\
EFPP(CVPR'2025) & 25.73 & 0.7771 & 22.60 \\
DiffProtect (PR'26) & 25.10 & 0.8262 & 25.23 \\
Flux-Guard (Ours) & 25.19 & 0.8263 & 24.71 \\
\hline
\end{tabular}
\label{table3}
\end{table}

\begin{figure}[t]
    \centering
     \includegraphics[width=\linewidth, trim=0.2cm 0.5cm 0.2cm 0.2cm, clip]{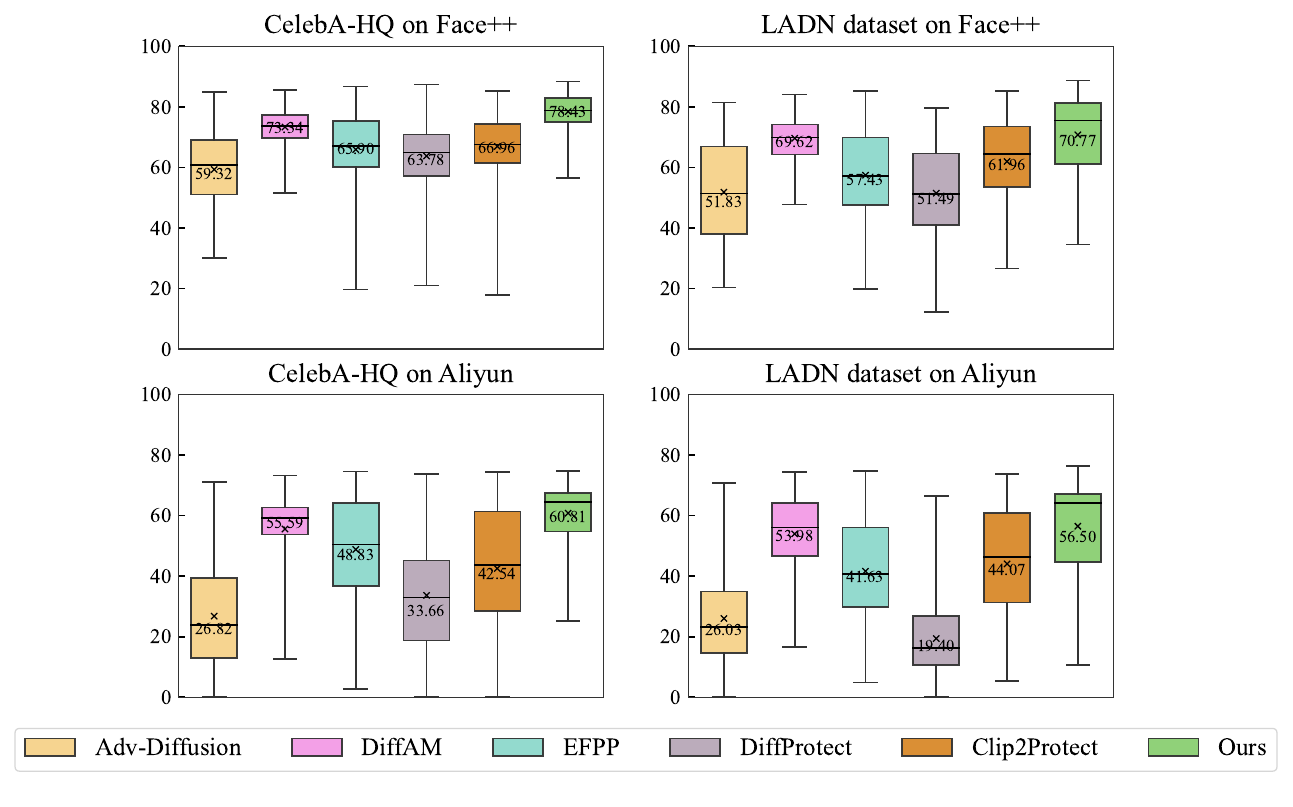}
    \caption{The confidence scores (↑) returned from commercial APIs, Face++ and Aliyun. Flux-Guard has higher and more stable confidence scores than other facial privacy protection methods.}
    \label{fig4}
\end{figure}

Image Quality. We selected five state-of-the-art methods from the past three years as benchmarks for comparison. Fig.\ref{fig3} illustrates the visual quality comparison of generated protective images across different methods. The proposed method maintains high confidence scores under both commercial APIs. In terms of visual quality, Clip2protect and DiffAM preserve basic makeup capabilities, while Adv-Diffusion, DiffProtect, and EFPP exhibit relatively weak facial modifications. In contrast, Flux-Guard delivers prompt-aligned edits while maintaining higher adversarial strength, better meeting customized editing needs, and producing more natural-looking protected images. Table \ref{table3} provides a quantitative evaluation using SSIM, PSNR, and FID metrics. Our scheme consistently achieves higher SSIM scores, while our method performs additional editing operations on faces rather than preserving semantic invariance, resulting in PSNR and FID metrics that underperform compared to DiffProtect. Overall, this approach generates images that are relatively natural while meeting customized editing requirements.

\begin{figure}[t]
    \centering
    \begin{minipage}[b]{0.48\columnwidth}
        \centering
        \includegraphics[width=\linewidth, trim=0.1cm 0.5cm 0.1cm 0.2cm, clip]{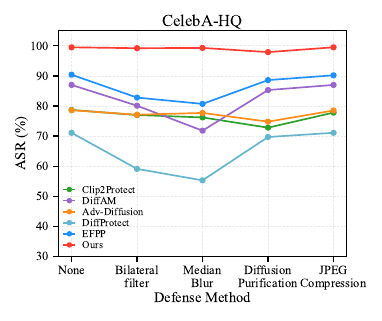}
    \end{minipage}
    \hfill
    \begin{minipage}[b]{0.48\columnwidth}
        \centering
        \includegraphics[width=\linewidth, trim=0.1cm 0.5cm 0.1cm 0.2cm, clip]{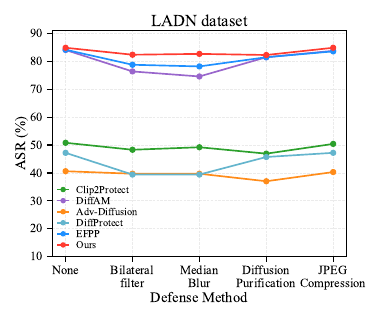}
    \end{minipage}
    \caption{ASR(\%) under various defense methods. We test various facial privacy protection methods under Bilateral filtering, JPEG compression, Median Blur and
    Diffusion Purification. Flux-Guard maintains the highest ASR by large margins.}
    \label{fig5}
\end{figure}

\subsubsection{Attack Performance on Commercial APIs}
To further evaluate real-world scenarios for facial privacy protection, we conducted comparative experiments using two commercial APIs. Here, we selected the widely used Face++ and Aliyun as FR models and evaluated multiple attack methods on the CelebA-HQ and LADN datasets. Figure 4 shows the quantification results of two commercial APIs during facial verification. The evaluation measures the confidence score between the protected image and the target identity. A higher confidence score indicates a greater similarity between the two images, which translates to stronger target impersonation. The values in Fig.\ref{fig4} represent the average confidence scores returned. The results show that our method is effective on both APIs, consistently achieving high average confidence scores, and demonstrates stable and effective attack performance across both datasets. These findings indicate that, compared to other state-of-the-art methods, our method possesses greater attack potential in black-box scenarios involving unknown mechanisms and can effectively target real-world applications.
\begin{figure}[!t]
    \centering
     \includegraphics[width=\linewidth, trim=7.2cm 6cm 14.2cm 5.4cm, clip]{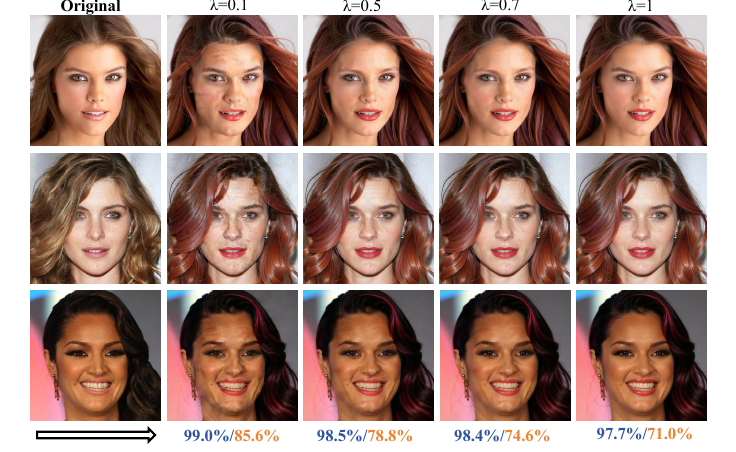}
    \caption{Generated images for different values of $\lambda$. The labels below the images show the ASR scores for face verification and the R1-T scores for face identification on the CelebA-HQ dataset (blue/yellow) for each value.}
    \label{fig6}
\end{figure}
\subsubsection{Comparison Against Adversarial Defense Methods}
To further validate the robustness of the Flux-Guard, this section evaluates its effectiveness under four typical defense mechanisms, specifically bilateral filtering, JPEG compression, median blurring, and diffusion-based defense DiffPure \cite{ref55}. These correspond to different types of defensive perspectives, thereby enabling an assessment of how well the protective images generated by these methods retain their performance following post-processing or purification operations. Fig.\ref{fig5} presents the experimental results for the two datasets. Overall, across all defense settings, our method maintains a high attack success rate, and its performance consistently outperforms that of other comparison methods. This indicates that the proposed method can generate high-quality protective images and effectively withstand common defense and purification operations, thereby demonstrating superior robustness.

\subsubsection{Ablation Study}
We evaluate the attack success rate (ASR) for face verification and identification under different values of $\lambda$, with visual results shown in Fig. \ref{fig6}. The results indicate that increasing $\lambda$ improves the visual quality of the generated images, but reduces both ASR and R1-T scores, as a larger $\lambda$ imposes stricter constraints on the perceptual structure of the images, limiting the magnitude and effectiveness of adversarial perturbations. To further validate the effectiveness of the perceptual-loss-driven adaptive adversarial weighting strategy, we performed ablation experiments, with the results summarized in Table \ref{table4}. Without this adaptive mechanism, attack success rates drop, indicating that dynamically adjusting the weights of identity injection can significantly improve attack performance.
\begin{table}[htbp]
\centering
\footnotesize
\setlength{\tabcolsep}{7.2pt}
\renewcommand{\arraystretch}{0.7}
\caption{Impact of the adaptive adversarial weighting strategy driven by perceptual loss in the face verification and identification.}
\begin{tabular}{lcccccccc}
\toprule
\multicolumn{9}{c}{Face Verification} \\
\midrule
 & \multicolumn{2}{c}{IR152} & \multicolumn{2}{c}{IRSE50} & \multicolumn{2}{c}{FaceNet} & \multicolumn{2}{c}{MobileFace} \\
\cmidrule(lr){2-9}
W/o adaptive adversarial & \multicolumn{2}{c}{75.5} & \multicolumn{2}{c}{94.8} & \multicolumn{2}{c}{46.5} & \multicolumn{2}{c}{92.9} \\
W/ adaptive adversarial & \multicolumn{2}{c}{85.4} & \multicolumn{2}{c}{95.8} & \multicolumn{2}{c}{62.9} & \multicolumn{2}{c}{99.5} \\
\midrule
\multicolumn{9}{c}{Face Identification} \\
\midrule
 & \multicolumn{2}{c}{IR152} & \multicolumn{2}{c}{IRSE50} & \multicolumn{2}{c}{FaceNet} & \multicolumn{2}{c}{MobileFace} \\
 & R1-T & R5-T & R1-T & R5-T & R1-T & R5-T & R1-T & R5-T \\
\cmidrule(lr){2-9}
W/o adaptive adversarial & 28.6 & 70.2 & 53.6 & 93.8 & 12.4 & 58.2 & 28.8 & 72.2 \\
W/ adaptive adversarial & 46.6 & 82.2 & 78.8 & 98.9 & 26.3 & 67.8 & 82.4 & 95.4 \\
\bottomrule
\end{tabular}
\label{table4}
\end{table}
\section{Conclusion}
In this work, we introduce Flux-Guard, a novel face editing framework with facial adversarial attacks.  Flux-Guard enables local facial region manipulation within a unified generative pipeline, simultaneously injecting target identity and applying adaptive adversarial perturbations to ensure effective facial privacy protection. Experimental results demonstrate that Flux-Guard outperforms existing methods by satisfying the need for facial editing while effectively addressing black-box attacks against diverse face recognition models and commercial APIs. In future work, we plan to explore this facial privacy protection paradigm for more fine-grained and diverse facial manipulation tasks, aiming to protect facial privacy in the era of artificial intelligence.
\section{Acknowledgment}
This work is supported by the National Science Foundation of China under Grant U21B2020 and the 111 Project under Grant B21049.
\section{CRediT authorship contribution statement}
\textbf{Jie Wang}: Writing - review \& editing, Writing – original draft, Visualization, Validation, Supervision, Software, Resources, Project administration, Methodology, Investigation, Formal analysis, Data curation, Conceptualization. \textbf{Ru Zhang}: Formal analysis, Project administration, Funding acquisition, Investigation, Supervision. \textbf{Jianyi Liu}: Formal analysis, Visualization, Software, Supervision, Conceptualization


\begin{thebibliography}{00}

%% For numbered reference style

\bibitem{ref1}
 E.Wenger, S. Shan, H. Zheng, B. Y. Zhao,
  Sok: Anti-facial recognition technology,
 IEEE Symposium on Security and Privacy,
 2023,
 pp.864-881.

\bibitem{ref2}
B. Meden, P. Rot, P. Terhörst, N. Damer, A. Kuijper, et al,
  Privacy–Enhancing Face Biometrics: A Comprehensive Survey,
   IEEE Transactions on Information Forensics and Security,
   2021,
   pp. 4147-4183.
\bibitem{ref3}
C. Chen, M. Sun, X. Gong, Y. Chen, Q. Wang,
 A Survey on Facial Image Privacy Preservation in Cloud-Based Services,
   arXiv preprint arXiv,
   2025.
  \bibitem{ref4}
L. Laishram, M. Shaheryar, J. T. Lee, S. K. Jung,
  Toward a Privacy-Preserving Face Recognition System: A Survey of Leakages and Solutions,
   ACM Computing Surveys,
   2025.

    \bibitem{ref5}
I. Goodfellow, J. Shlens, C. Szegedy,
  Explaining and Harnessing Adversarial Examples,
   International Conference on Learning Representations(ICLR),
  2014.

    \bibitem{ref6}
A. Madry, A. Makelov, L. Schmidt, D. Tsipras, A. Vladu,
  Towards deep learning
models resistant to adversarial attacks,
   International Conference on Learning Representations(ICLR),
  2018.

    \bibitem{ref7}
Y. Dong, T. Pang, H. Su, J. Zhu,
  Evading defenses to transferable adversarial examples by translation-invariant attacks,
   IEEE/CVF Conference on Computer Vision and Pattern Recognition (CVPR),
2019,
   pp. 4312-4321.

    \bibitem{ref8}
X. Yang, Y. Dong, T. Pang, H. Su, J.Zhu, et al,
  Towards face encryption by generating adversarial identity masks,
   International Conference on Computer Vision(ICCV),
   2021,
   pp. 3897-3907.


    \bibitem{ref9}
F. Zhou, Q. Zhou, H. Ling, X. Lu,
 Adversarial Attacks on Both Face Recognition and Face Anti-spoofing Models,
   International Joint Conference on Artificial Intelligence,
   2025,
  pp. 2494-2502.

     \bibitem{ref10}
B. Yin, W. Wang, T. Yao, J. Guo, Z. Kong, et al,
  Adv-Makeup: A New Imperceptible and Transferable Attack on Face Recognition,
   International Joint Conference on Artificial Intelligence,
   2021,
   pp. 1252-1258.

     \bibitem{ref11}
S. Hu, X. Liu, Y. Zhang, M. Li, L. Y. Zhang, et al,
Protecting Facial Privacy: Generating Adversarial Identity Masks via Style-robust Makeup Transfer,
   IEEE/CVF Conference on Computer Vision and Pattern Recognition (CVPR),
   2022,
   pp. 14994-15003.

     \bibitem{ref12}
F. Shamshad, M. Naseer, K. Nandakumar,
  CLIP2Protect: Protecting Facial Privacy Using Text-Guided Makeup via Adversarial Latent Search,
   IEEE/CVF Conference on Computer Vision and Pattern Recognition (CVPR),
   2023,
   pp. 20595-20605.


     \bibitem{ref13}
F. Shamshad, M. Naseer, K. Nandakumar,
  Makeup-Guided Facial Privacy Protection via Untrained Neural Network Priors,
   European Conference on Computer Vision Workshops (ECCV),
 2024.
     \bibitem{ref14}
M. Li, J. Wang, H. Zhang, Z. Zhou, S. Hu, et al,
  Transferable adversarial facial images for privacy protection,
    ACM International Conference on Multimedia (ACM MM),
    2024,
    pp. 10649-10658.

     \bibitem{ref15}
Y. Lyu, Y. Jiang, Z. He, B. Peng, Y. Liu, et al,
  3D-Aware Adversarial Makeup Generation for Facial Privacy Protection,
    IEEE Transactions on Pattern Analysis and Machine Intelligence,
    2023,
    pp. 13438-13453.

     \bibitem{ref16}
Y. Sun, L. Yu, H. Xie, J. Li, Y. Zhang,
  DiffAM: Diffusion-Based Adversarial Makeup Transfer for Facial Privacy Protection,
    IEEE/CVF Conference on Computer Vision and Pattern Recognition (CVPR),
    2024,
    24584-24594.

    \bibitem{ref17}
Z. Wang, T. Pang, C. Du, M. Lin, W. Liu, et al,
  Better Diffusion Models Further Improve Adversarial Training,
    International Conference on Machine Learning (ICML),
    2023,
    pp. 36246-36263.


    \bibitem{ref18}
W. Nie, B. Guo, Y. Huang, C. Xiao, A. Vahdat, et al,
  Diffusion Models for Adversarial Purification,
    International Conference on Machine Learning (ICML),
 2022.

    \bibitem{ref19}
N. Carlini, F.Tramer, K. D. Dvijotham, L. Rice, M. Sun, et al,
 (Certified!!) Adversarial Robustness for Free! ,
    International Conference on Learning Representations(ICLR),
 2023.

    \bibitem{ref20}
J. Chen, H. Chen, K. Chen, Y. Zhang, Z. Zou, et al,
 Diffusion Models for Imperceptible and Transferable Adversarial Attack,
    IEEE Transactions on Pattern Analysis and Machine Intelligence,
    2025,
    pp. 961-977.
    \bibitem{ref21}
H. Xue, A. Araujo, B. Hu, Y. Chen,
  Diffusion-Based Adversarial Sample Generation for Improved Stealthiness and Controllability,
    Annual Conference on Neural Information Processing Systems(NeurIPS),
 2023.

    \bibitem{ref22}
D. Liu, X. Wang, C. Peng, N. Wang, R. Hu, et al,
  Adv-Diffusion: Imperceptible Adversarial Face Identity Attack via Latent Diffusion Model,
   Proceedings of the AAAI Conference on Artificial Intelligence(AAAI),
   2024,
   pp. 3585-3593.

 
    \bibitem{ref23}
J. An, W. Zhang, D. Wu, Z. Lin, J. Gu, et al,
  SD4Privacy: Exploiting Stable Diffusion for Protecting Facial Privacy,
     IEEE International Conference on Multimedia and Expo (ICME),
 2024.
    \bibitem{ref24}
C. Hu, Y. Li, Z. Feng, X. Wu,
 Toward Transferable Attack via Adversarial Diffusion in Face Recognition,
      IEEE Transactions on Information Forensics and Security,
      2024,
    pp. 5506-5519.
    \bibitem{ref25}
J. Liu, C. P. Lau, Z. Guo, Y. Guo, Z. Wang, et al,
  DiffProtect: Generative adversarial examples using diffusion models for facial privacy protection,
      Pattern Recognition,
       2026,
    pp. 112780.

    \bibitem{ref26}
L. Wang, Q. Hu, W. Lu, X. Luo,
  Diffusion-based Adversarial Identity Manipulation for Facial Privacy Protection,
      ACM International Conference on Multimedia (ACM MM),
 2025.
  \bibitem{ref27}
D. Han, S. Mohamed, Y. Li, J. Denzler,
 Diffusion-based Identity-Preserving Facial Privacy Protection,
     IEEE International Conference on Acoustics, Speech and Signal Processing (ICASSP),
 2025.
  \bibitem{ref28}
J. Li, J. Dong, J. Lai, X. Xie,
  Imperceptible diffusion modification for facial privacy protection,
     Neurocomputing,
     2025,
   pp. 130614.

    \bibitem{ref29}
A. Salar, Q. Liu, Y. Tian, G. Zhao,
 Enhancing Facial Privacy Protection via Weakening Diffusion Purification,
     IEEE/CVF Conference on Computer Vision and Pattern Recognition (CVPR),
     2025,
   pp. 8235-8244

    \bibitem{ref30}
Y. Li, C. Hu, X. Wu,
  Transferable Stealthy Adversarial Example Generation via Dual-Latent Adaptive Diffusion for Facial Privacy Protection,
     IEEE Transactions on Information Forensics and Security,
     2025,
   pp. 9427-9440.

    \bibitem{ref31}
Y. Huang, L. Xie, X. Wang, Z. Yuan, X. Cun, et al,
SmartEdit: Exploring Complex Instruction-Based Image Editing with Multimodal Large Language Models,
     IEEE/CVF Conference on Computer Vision and Pattern Recognition (CVPR),
     2024,
  pp. 8362-8371.


    \bibitem{ref32}
J. Wang, J. Pu, Z. Qi, J. Guo, Y. Ma, et al,
Taming Rectified Flow for Inversion and Editing,
     International Conference on Machine Learning (ICML),
 2025.

    \bibitem{ref33}
T. He, R. Wang, Y. Chen, D. Song, N. Chen, et al,
  Flux-Sculptor: Text-Driven Rich-Attribute Portrait Editing through Decomposed Spatial Flow Control,
    arXiv preprint arXiv,
 2025.

    \bibitem{ref34}
X. Liu, C. Gong, Q. liu,
  Flow Straight and Fast: Learning to Generate and Transfer Data with Rectified Flow,
     International Conference on Learning Representations(ICLR),
2023.

   \bibitem{ref35}
R. Rombach, A. Blattmann, D. Lorenz, P. Esser, B. Ommer,
High-Resolution Image Synthesis with Latent Diffusion Models,
     IEEE/CVF Conference on Computer Vision and Pattern Recognition (CVPR),
      2022,
  pp. 10674-10685.

   \bibitem{ref36}
J. Song, C. Meng, S. Ermon,
  Denoising Diffusion Implicit Models,
       International Conference on Learning Representations(ICLR),
 2021.

   \bibitem{ref37}
R. Gal, Y. Alaluf, Y. Atzmon, O. Patashnik, A. Bermano, G. Chechik, et al,
 An Image is Worth One Word: Personalizing Text-to-Image Generation using Textual Inversion,
     International Conference on Learning Representations(ICLR),
2023.

    \bibitem{ref38}
Z. Guo, Y. Wu, Z. Chen, L. Chen, P. Zhang, et al,
  PuLID: Pure and Lightning ID Customization via Contrastive Alignment,
    Annual Conference on Neural Information Processing Systems(NeurIPS),
 2024.

  \bibitem{ref39}
L. Kong, K. Wu, X. Hu, W. Han, J. Peng, et al,
  AnyMaker: Zero-shot General Object Customization via Decoupled Dual-Level ID Injection,
   IEEE/CVF Conference on Computer Vision and Pattern Recognition (CVPR),
 2025.


  \bibitem{ref40}
X. Peng, J. Zhu, B. Jiang, Y. Tai, D. Luo, et al,
  PortraitBooth: A Versatile Portrait Model for Fast Identity-preserved Personalization,
   IEEE/CVF Conference on Computer Vision and Pattern Recognition (CVPR),
 2024.


  \bibitem{ref41}
D. Podell, Z. English, K.Lacey, A. Blattmann, T. Dockhorn, et al,
  SDXL: Improving Latent Diffusion Models for High-Resolution Image Synthesis,
      International Conference on Learning Representations(ICLR),
 2024.

   \bibitem{ref42}
P. Esser, S. Kulal, A. Blattmann, R. Entezari, J. Müller, et al,
 Scaling Rectified Flow Transformers for High-Resolution Image Synthesis,
     International Conference on Machine Learning (ICML),
 2024.

   \bibitem{ref43}
 Black Forest Labs,
  Flux. https://github.com/black-forest-labs/flux,
 2024.
    \bibitem{ref44}
Y. Lipman, R. T. Q. Chen, H. Ben-Hamu, M. Nickel, M. Le,
  Flow Matching for Generative Modeling,
      International Conference on Learning Representations(ICLR),
 2023.

    \bibitem{ref45}
D. P Kingma, M. Welling,
Auto-encoding variational bayes,
     arXiv preprint arXiv,
 2013.

    \bibitem{ref46}
W. Peebles, S. Xie,
  Scalable diffusion models with transformers,
    International Conference on Computer Vision(ICCV),
 2023.


    \bibitem{ref47}
J. Ho, A. N. Jain, P. Abbeel,
  Denoising Diffusion Probabilistic Models,
    Annual Conference on Neural Information Processing Systems(NeurIPS),
 2020.

    \bibitem{ref48}
T. Karras, T. Aila, S. Laine, J. Lehtinen,
 Progressive Growing of GANs for Improved Quality, Stability, and Variation,
    International Conference on Learning Representations(ICLR),
 2018.

    \bibitem{ref49}
Q. Gu, G. Wang, M. T. Chiu, Y. Tai, C. Tang,
  Ladn: Local adversarial disentangling network for facial makeup and de-makeup,
     International Conference on Computer Vision(ICCV),
     2019,
    pp. 10481-10490.

    \bibitem{ref50}
J. Deng, J. Guo, N. Xue, S. Zafeiriou,
  Arcface: Additive angular margin loss for deep face recognition,
      IEEE/CVF Conference on Computer Vision and Pattern Recognition (CVPR),
      2019,
     pp. 4690-4699.
 
    \bibitem{ref51}
J.Hu, L. Shen, G. Sun,
  Squeeze-and-excitation networks,
     IEEE/CVF Conference on Computer Vision and Pattern Recognition (CVPR),
     2018,
     pp. 7132-7141.
    \bibitem{ref52}
F. Schroff, D. Kalenichenko, J. Philbin,
  Facenet: A unified embedding for face recognition and clustering,
     IEEE/CVF Conference on Computer Vision and Pattern Recognition (CVPR),
     2015,
     pp. 815-823.

    \bibitem{ref53}
 S. Chen, Y. Liu, X. Gao, Z. Han,
 MobileFaceNets: Efficient CNNs
for accurate real-time face verification on mobile devices,
     Chinese Conference on Biometric Recognition,
     2018,
     pp. 428-438.

    \bibitem{ref54}
M. Heusel, H. Ramsauer, T. Unterthiner, B. Nessler, S. Hochreiter,
  GANs trained by a two time-scale update rule converge to a local nash equilibrium,
       Annual Conference on Neural Information Processing Systems(NeurIPS),
       2017,
     pp. 6629-6640.


  \bibitem{ref55}
W. Nie, B. Guo, Y. Huang, C. Xiao, A. Vahdat, et al,
  Diffusion models for adversarial purification,
   International Conference on Machine Learning (ICML),
 2022.



 
\end{thebibliography}
\end{document}